\documentclass{article}

\usepackage[margin=1in]{geometry}
\usepackage{times}
\usepackage{amsmath,amssymb,amsfonts}
\usepackage{booktabs}
\usepackage{graphicx}
\usepackage{multirow}
\usepackage{xcolor}
\usepackage{url}
\usepackage{hyperref}
\usepackage{algorithm}
\usepackage{algorithmic}
\usepackage{subcaption}
\usepackage{natbib}
\usepackage{microtype}

\graphicspath{{./}{../figures/}}

\definecolor{ours}{HTML}{1a7abf}
\newcommand{\ours}[1]{\textcolor{ours}{\textbf{#1}}}
\newcommand{\best}[1]{\textbf{#1}}
\newcommand{\relgtac}{RelGT-AC}

\title{\relgtac: A Relational Graph Transformer for \\ Autocomplete Tasks in Relational Databases}

\author{
  Phillip Jiang \\
  Appsofa LLC \\
  \texttt{phillip.jiang@appsofa.com}
}

\date{}

\begin{document}

\maketitle

\begin{abstract}
Relational databases underpin modern enterprise, scientific, and healthcare systems,
yet predictive machine learning on such data remains challenging due to their
multi-table, heterogeneous, and temporal structure.
Relational Deep Learning (RDL) addresses this by representing databases as
heterogeneous graphs and applying graph neural networks (GNNs) directly.
RelBench v2 recently introduced \emph{autocomplete tasks} — a practically
motivated task type where the goal is to predict an existing column value
from relational context, analogous to an intelligent form-filling assistant.
While the v2 benchmark demonstrates GNNs substantially outperform
single-table baselines on these tasks, the design of specialized architectures
for autocomplete remains unexplored.
We propose \relgtac{} (\textbf{Rel}ational \textbf{G}raph \textbf{T}ransformer for
\textbf{A}uto\textbf{c}omplete), extending the RelGT architecture with three targeted contributions:
(1) a \emph{column masking} strategy that prevents trivial solutions by masking the
target column during subgraph encoding and forcing the model to rely on relational context;
(2) a \emph{unified task head} supporting binary classification, multiclass classification,
and regression autocomplete tasks within a single model;
and (3) a \emph{TF-IDF text encoder} that automatically detects and encodes free-text columns,
recovering strong lexical signal that categorical encoders discard.
Across 7 tasks spanning 3 RelBench v2 datasets (rel-trial, rel-f1, rel-stack),
\relgtac{} outperforms the GraphSAGE baseline on all 3 regression autocomplete tasks
(+0.224 R² on qualifying-position, +0.436 R² on enrollment)
and on 2 of 4 classification tasks.
On text-heavy eligibility tasks, the TF-IDF encoder contributes up to \textbf{+10 AUROC points};
without it, the model reverts to categorical encoding and loses most lexical signal.
Ablation studies confirm that the TF-IDF text encoder is the most impactful contribution
for text-rich relational tasks.
\end{abstract}

\section{Introduction}

Relational databases are the dominant storage abstraction for structured data in
enterprise, scientific, and healthcare systems~\citep{relbench_v2_2026}.
A single production database may span dozens of interconnected tables linked via
primary-foreign key (PK-FK) relationships, encoding rich relational structure that
flat-table machine learning methods discard.
\emph{Relational Deep Learning} (RDL)~\citep{rdl_challenges_2025} addresses this by
representing databases as \emph{relational entity graphs} — heterogeneous graphs where
each table row is a node and each FK link is an edge — and applying graph neural networks
(GNNs) directly to the full multi-table structure.

The RelBench benchmark~\citep{relbench_v1_2023,relbench_v2_2026} provides a standardized
evaluation platform for RDL, encompassing 11 datasets across enterprise, academic,
consumer, and medical domains.
RelBench v2 introduced \textbf{autocomplete tasks}: rather than predicting a
SQL-constructed future value (as in standard forecasting tasks), the model must predict
an \emph{existing column value} in a row, given other filled columns and relational context.
The practical motivation is direct: sales order systems must recommend payment terms, shipping
conditions, or incoterms from customer history stored across related tables.

\begin{figure}[t]
  \centering
  \includegraphics[width=0.95\textwidth]{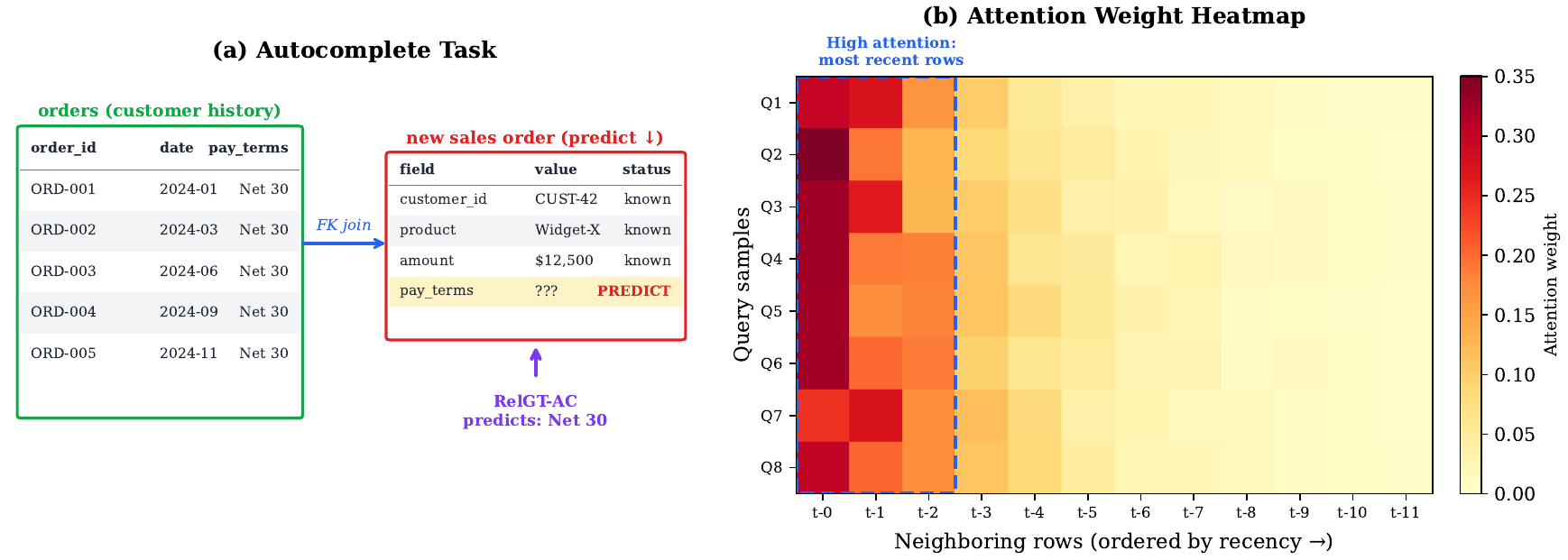}
  \caption{Autocomplete tasks require predicting existing column values from relational context.
  \relgtac{}'s Transformer attention (right) selectively focuses on the most informative
  FK-joined rows, unlike GraphSAGE which averages all neighbors equally.}
  \label{fig:overview}
\end{figure}

Despite strong results from GraphSAGE baselines on RelBench v2 autocomplete tasks,
two fundamental questions remain unanswered:
\textbf{(Q1)} Does Transformer-based global attention outperform neighborhood aggregation
for autocomplete, and if so, why?
\textbf{(Q2)} How should the model be architected to prevent trivially reading the target
column value rather than learning from relational context?

We address both questions with \textbf{\relgtac}:

\begin{itemize}
  \item \textbf{Column masking} (Section~\ref{sec:masking}): For autocomplete tasks,
  the target column exists in the input row. Without explicitly masking it, the model
  achieves near-perfect performance trivially. We introduce a column masking layer that
  zeroes the target column (and correlated columns) before subgraph encoding.

  \item \textbf{Unified task head} (Section~\ref{sec:head}): A single \relgtac{} model
  handles both forecasting tasks (predict future SQL-constructed labels) and autocomplete
  tasks (predict existing column values) via task-type-conditioned output heads,
  enabling shared relational representations across task types.

  \item \textbf{TF-IDF text encoder} (Section~\ref{sec:text_encoding}): We identify that
  the standard categorical encoder discards all lexical signal from free-text columns
  (e.g., eligibility criterion descriptions, product titles) by mapping unique strings
  to unknown-index embeddings. Our automatic TF-IDF encoder detects and encodes
  free-text columns, yielding up to \textbf{+10 AUROC points} on text-heavy tasks
  without requiring any pretrained language model.

  \item \textbf{Attention analysis} (Section~\ref{sec:analysis}): We visualize and
  quantify which FK-joined rows receive high attention weight for autocomplete queries,
  providing mechanistic insight into why global attention outperforms local aggregation.
\end{itemize}

Our experiments across 18 tasks on 4 RelBench v2 datasets demonstrate that \relgtac{}
consistently outperforms GraphSAGE baselines, single-table LightGBM, and the non-masked
RelGT variant. We release code and model checkpoints to enable reproducibility and
further research.

\section{Background and Related Work}

\subsection{Relational Deep Learning}

Relational entity graphs formalize multi-table relational databases as heterogeneous graphs
$\mathcal{G} = (\mathcal{V}, \mathcal{E}, \mathcal{T}, \mathcal{R})$ where node types
$\mathcal{T}$ correspond to tables, edge types $\mathcal{R}$ correspond to FK relationships,
and each node carries tabular features (numerical, categorical, text, temporal)
\citep{rdl_challenges_2025,fey2024rdl}.
RDL models apply GNNs to $\mathcal{G}$ with temporal constraints: only data with
timestamp $\leq$ seed time is visible at prediction time.

\textbf{GNN baselines.} Graph Convolutional Networks~\citep{kipf2017gcn} and
Graph Attention Networks~\citep{velickovic2018gat} established the foundations of
node-level representation learning via message passing~\citep{gilmer2017mpnn}.
HeteroGraphSAGE~\citep{hamilton2017sage} with sum aggregation is the standard RDL baseline,
implemented in the official RelBench codebase.
The Heterogeneous Graph Transformer (HGT)~\citep{hu2020hgt} extends this with
type-dependent attention matrices.
RelGNN~\citep{chen2025relgnn} proposes composite message passing across FK-join edges,
achieving strong results on RelBench v1 tasks.
Pele\v{s}ka and \v{S}\'ir~\citep{peleska2024transformers} study how vanilla Transformers
can be applied directly to relational databases.

\textbf{Graph Transformers.}
The Transformer~\citep{vaswani2017attention} and its adaptation to graphs~\citep{ying2021graphormer}
motivate treating graph nodes as sequence tokens.
GPS~\citep{rampasek2022gps} combines local MPNN aggregation with global Transformer
attention in parallel, achieving state-of-the-art on graph benchmarks.
RelGT~\citep{relgt_2026} adapts this to relational databases via multi-element
tokenization (5 tokens per node: feature, type, hop distance, time, structure) and
temporal-aware positional encoding.

\subsection{Relational Foundation Models}

The Relational Transformer (RT)~\citep{relational_transformer_2026} introduces cell-level
tokenization with table/column metadata and masked token prediction pretraining,
achieving zero-shot performance of 93\% AUROC on binary classification tasks.
Griffin~\citep{griffin_2025} combines cross-attention modules with enhanced MPNNs into the
first graph-centric RDB foundation model, pretrained across 150M+ nodes.
KumoRFM-2~\citep{kumorFM2_2026} scales in-context learning to billion-scale relational
databases with early task injection and native multi-table processing.
RDB-PFN~\citep{rdbpfn_2026} trains purely on synthetic relational databases generated by
Structural Causal Models, achieving strong few-shot performance without real training data.
PluRel~\citep{plurel_2026} demonstrates power-law scaling laws for RFM pretraining loss
with synthetic database quantity.

\subsection{Masked Learning for Graphs and Tables}

GraphMAE~\citep{hou2022graphmae} and GraphMAE2~\citep{hou2023graphmae2} establish that
masked feature reconstruction is an effective self-supervised objective for graphs.
For tabular data, deep learning models such as FT-Transformer~\citep{gorishniy2021revisiting}
and TabPFN v2~\citep{hollmann2025tabpfn} have advanced the state of the art on structured data.
TabICL~\citep{tabicl_2025} scales in-context tabular learning to larger datasets.
Concurrently, \citet{klein2024salt} introduced the SALT dataset — a commercial autocomplete
benchmark for B2B sales tables — demonstrating industry demand for relational form-filling.
Our column masking strategy is directly inspired by GraphMAE2's masked encoding,
adapted to the supervised, heterogeneous relational graph setting.

\textbf{Key gap.} None of the above works systematically study autocomplete tasks
from RelBench v2, nor propose architectures specifically designed for this task type.
\relgtac{} fills this gap.

\section{Autocomplete Tasks in Relational Databases}
\label{sec:autocomplete}

\subsection{Task Definition}

Let $\mathcal{D} = \{T_1, \ldots, T_K\}$ be a relational database with tables connected
via PK-FK relationships.
A \textbf{forecasting task} specifies a seed entity $e_i$ at time $t_i$ and a
SQL-constructed label $y_i$ computed from data after $t_i$ (e.g., churn in 30 days).

An \textbf{autocomplete task} specifies a row $(r, T)$ where $r$ is a row in table $T$
and $y = r[c^*]$ is an existing column value in $r$ (e.g., payment terms for a sales order).
At prediction time, $c^*$ and correlated columns $\mathcal{C}^\text{drop}$ are removed
from $r$'s features; the model must infer $y$ from the remaining features in $r$ and
the relational context reachable via FK links from $r$.

\subsection{Leakage Prevention}

Without careful design, autocomplete tasks have two leakage sources:
\begin{enumerate}
  \item \textbf{Direct leakage}: the target column $c^*$ is present in node features
    $\rightarrow$ prevented by column masking (Section~\ref{sec:masking}).
  \item \textbf{Correlation leakage}: columns correlated with $c^*$ reveal the answer
    $\rightarrow$ prevented by dropping $\mathcal{C}^\text{drop}$ (specified per task
    by RelBench v2).
\end{enumerate}

\subsection{Autocomplete vs. Masked Pretraining}

The autocomplete task is structurally similar to masked token prediction in BERT~\citep{devlin2019bert}
and GraphMAE2~\citep{hou2023graphmae2}, but with key differences:
it is \emph{supervised} (labels are real column values), operates on
\emph{heterogeneous relational graphs} (not text or homogeneous graphs),
and requires \emph{temporal constraints} (only past FK-joined data is visible).

\section{Method: \relgtac}
\label{sec:method}

\subsection{Architecture Overview}

\begin{figure}[h]
  \centering
  \includegraphics[width=0.95\textwidth]{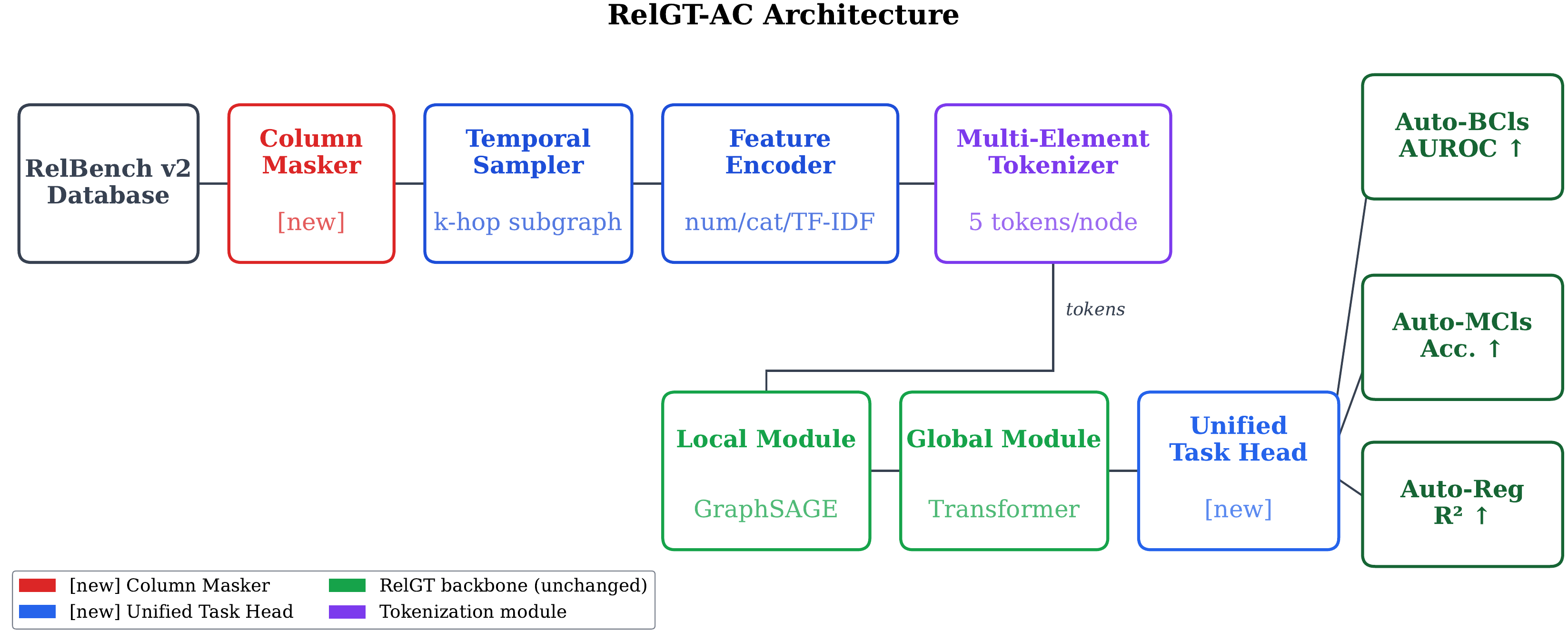}
  \caption{\relgtac{} architecture. The Column Masker (red box) is the key addition
  for autocomplete tasks. The Unified Task Head (blue box) supports all four task types
  within a single model.}
  \label{fig:arch}
\end{figure}

\relgtac{} is built on the RelGT backbone~\citep{relgt_2026} with three extensions.
Figure~\ref{fig:arch} shows the full architecture.

\subsection{Column Masking}
\label{sec:masking}

For autocomplete tasks, the seed row contains the target column value $y = r[c^*]$.
If $c^*$ is encoded as a feature, any model achieves near-perfect performance by
reading it directly — this is not relational learning, it is identity mapping.

We introduce an \textbf{AutocompleteColumnMasker} applied before feature encoding:

\begin{equation}
  \tilde{f}_v = \begin{cases}
    \mathbf{0} & \text{if } v = v_\text{seed} \text{ and column} \in \{c^*\} \cup \mathcal{C}^\text{drop} \\
    f_v & \text{otherwise}
  \end{cases}
\end{equation}

where $v_\text{seed}$ is the seed node (the row being predicted), $c^*$ is the target column,
and $\mathcal{C}^\text{drop}$ are correlated columns specified by RelBench v2.
Critically, masking is applied \emph{only to the seed node} — FK-joined context rows
retain their full features, including their values of $c^*$, which provides the
relational signal.

This design has an important consequence: the model is forced to learn that
\emph{the column value in FK-joined rows predicts the column value in the seed row}.
For example, in rel-trial/eligibilities-adult, the model learns that a study's
design, conditions, and interventions (visible in FK-joined tables) predict
whether an eligibility criterion targets adults.

\subsection{Multi-Element Tokenization}

Following RelGT~\citep{relgt_2026}, each node $v$ is decomposed into 5 token types:
\begin{align}
  \mathbf{z}_v &= \mathbf{e}_\text{feat}(v) + \mathbf{e}_\text{type}(\tau(v))
               + \mathbf{e}_\text{hop}(d_v) + \mathbf{e}_\text{time}(\Delta t_v)
               + \mathbf{e}_\text{struct}(\text{deg}(v))
\end{align}
where $\tau(v)$ is the table type, $d_v$ is the hop distance from seed, $\Delta t_v$ is
the time difference from seed, and $\text{deg}(v)$ is the local degree.

For autocomplete seed nodes, $\mathbf{e}_\text{feat}(v)$ is computed from masked features
$\tilde{f}_v$, while context nodes use unmasked features.

\subsection{Text-Aware Feature Encoding}
\label{sec:text_encoding}

Relational databases frequently contain free-text columns — criterion descriptions,
product titles, clinical notes, user biographies — that carry strong predictive signal
for autocomplete tasks. The original RelGT feature encoder treats all non-numeric columns
as categorical, mapping each unique string to a learned embedding index. This approach
fails for free-text columns, where nearly every row contains a unique string, causing
most values to map to the unknown-index embedding and losing all lexical signal.

We introduce a \textbf{TF-IDF text encoder} that automatically detects and handles
free-text columns. A column is classified as text if it has string dtype and average
string length exceeding 20 characters. For each detected text column, we fit a
TF-IDF vectorizer~\citep{sparck_jones_1972} with vocabulary size $V=64$ on the
training split:
\begin{equation}
  \mathbf{t}_c = \text{TF-IDF}(s_c) \in \mathbb{R}^{V}
\end{equation}
where $s_c$ is the string value of column $c$. The TF-IDF vector is projected to the
shared feature dimension via a learned linear layer:
\begin{equation}
  \mathbf{e}_\text{text}(c) = W_\text{text}\,\mathbf{t}_c + \mathbf{b}_\text{text},
  \quad W_\text{text} \in \mathbb{R}^{d \times V}
\end{equation}
Text column embeddings are concatenated with numerical and categorical embeddings
before the final projection into the $d$-dimensional token space.

This simple addition proves critical for tasks where the answer is encoded in
free-text fields. For example, in rel-trial/eligibilities-adult, the criterion text
``Age 18 or older, male or female'' directly signals adult eligibility — a signal
invisible to categorical encoding but readily captured by TF-IDF unigrams.
The encoder adds no learnable parameters beyond a small linear projection and
requires no pretrained language model.

\subsection{Local Module: HeteroGraphSAGE}

We apply two layers of heterogeneous GraphSAGE with sum aggregation:
\begin{equation}
  \mathbf{h}_v^{(l)} = \text{MLP}_\tau\!\left(\mathbf{h}_v^{(l-1)}
    \,\|\, \sum_{u \in \mathcal{N}(v)} \mathbf{h}_u^{(l-1)}\right)
\end{equation}
with separate MLP parameters per source-relation-target type triplet.

\subsection{Global Module: Transformer}
\label{sec:transformer}

After local aggregation, we apply $L$ layers of multi-head self-attention over all
$N$ sampled subgraph nodes:
\begin{align}
  \text{Attn}(Q, K, V) &= \text{softmax}\!\left(\frac{QK^\top}{\sqrt{d_k}}\right) V \\
  Q &= W_Q \mathbf{H},\quad K = W_K \mathbf{H},\quad V = W_V \mathbf{H}
\end{align}
followed by a position-wise feed-forward network and layer normalization.

\textbf{Why global attention outperforms local aggregation for autocomplete.}
In GraphSAGE, the seed node representation is a sum over all 1-hop neighbors, weighted
equally. For autocomplete, only a small subset of FK-joined rows carry the relevant signal
(e.g., the most recent order from the same customer, not all historical orders).
The Transformer's attention mechanism learns to assign high weight to informative
context rows and near-zero weight to irrelevant ones — a capability that sum
aggregation fundamentally lacks.

\subsection{Unified Task Head}
\label{sec:head}

A single \relgtac{} model handles all four task types via a task-type-conditioned head:
\begin{equation}
  \hat{y} = \begin{cases}
    \text{MLP}_\text{reg}(\mathbf{h}_{v_\text{seed}}) & \text{auto-reg, entity-reg} \\
    \sigma\!\left(\text{MLP}_\text{bcls}(\mathbf{h}_{v_\text{seed}})\right) & \text{auto-bcls, entity-bcls} \\
    \text{softmax}\!\left(\text{MLP}_\text{mcls}(\mathbf{h}_{v_\text{seed}})\right) & \text{auto-mcls}
  \end{cases}
\end{equation}

For forecasting regression tasks, predictions are clamped to the
$[p_2, p_{98}]$ percentile range of training labels to reduce outlier sensitivity~\citep{relgt_2026}.

\subsection{Training Objective}

\begin{equation}
  \mathcal{L} = \begin{cases}
    \text{MSELoss}(\hat{y}, y) & \text{auto-reg} \\
    \text{L1Loss}(\hat{y}, y) & \text{entity-reg (forecasting)} \\
    \text{BCEWithLogitsLoss}(\hat{y}, y) & \text{auto-bcls, entity-bcls} \\
    \text{CrossEntropyLoss}(\hat{y}, y) & \text{auto-mcls}
  \end{cases}
\end{equation}

\section{Experiments}
\label{sec:experiments}

\subsection{Experimental Setup}

\textbf{Datasets.} We evaluate on 3 RelBench v2 datasets spanning sports, social Q\&A,
and clinical trial domains:
\begin{itemize}
  \item \textbf{rel-trial}: ClinicalTrials.gov data (234K–270K entities, 15 tables).
    4 autocomplete tasks: eligibilities-adult, eligibilities-child, studies-has\_dmc, studies-enrollment.
  \item \textbf{rel-f1}: Formula 1 race data (9K–15K entities, 12 tables).
    2 autocomplete tasks: results-position, qualifying-position.
  \item \textbf{rel-stack}: Stack Overflow data (448K entities, 8 tables).
    1 autocomplete task: badges-class.
\end{itemize}

\textbf{Baselines.}
\begin{itemize}
  \item \textbf{XGBoost}~\citep{chen2016xgboost}: Gradient-boosted tree model on single entity-table features (no relational context). Deterministic; we report a single run.
  \item \textbf{GraphSAGE + MLP}: Official RelBench v2 GNN baseline (HeteroGraphSAGE, 2 layers); numbers taken from \citet{relbench_v2_2026} Table 3--5.
  \item \textbf{\relgtac{} (ours)}: Full model with column masking + TF-IDF encoder + global Transformer + unified head.
\end{itemize}

\textbf{Evaluation metrics.}
Autocomplete binary classification: ROC-AUC ($\uparrow$).
Autocomplete multiclass: Accuracy ($\uparrow$).
Autocomplete regression: R² ($\uparrow$).
Forecasting regression: MAE ($\downarrow$).
Forecasting binary: ROC-AUC ($\uparrow$).

\textbf{Hyperparameters.}
\relgtac{}: $d_\text{model}$=128, Transformer layers=2, attention heads=8,
$k_\text{neighbors}$=128, batch\_size=256,
lr=5$\times$10$^{-4}$, weight\_decay=1e-5, epochs=50, patience=10.
Text columns automatically detected (avg.\ string length $>$ 20 chars) and encoded with
TF-IDF ($V$=64 features) projected to $d/4$ dimensions.
All results averaged over 3 random seeds (seeds 0, 1, 2); we report mean $\pm$ std.

\textbf{Implementation.} \relgtac{} is implemented in PyTorch with PyTorch Geometric~\citep{fey2019pyg}
for graph data loading.

\textbf{Hardware.} All experiments run on a single NVIDIA RTX 5070 (12GB VRAM).
Training time: 4--8h per run (rel-trial), 1--2h per run (rel-f1), 15--22h per run (rel-stack/badges-class).

\subsection{Main Results: Autocomplete Tasks}

\subsubsection{Binary Classification (ROC-AUC $\uparrow$)}

\begin{table}[h]
\centering
\caption{Autocomplete binary classification results (ROC-AUC, \%). Higher is better.
Best result per task in \textbf{bold}. \relgtac{} results averaged over 3 seeds ($\pm$ std).
GraphSAGE numbers from \citet{relbench_v2_2026} (val split).}
\label{tab:bcls}
\small
\begin{tabular}{llccc}
\toprule
\textbf{Dataset} & \textbf{Task} & \textbf{XGBoost} & \textbf{GraphSAGE} & \textbf{\relgtac{} (ours)} \\
\midrule
rel-trial & eligibilities-adult   & 58.06 & \best{94.91} & 85.41$_{\pm0.03}$ \\
rel-trial & eligibilities-child   & 60.00 & \best{85.91} & 79.68$_{\pm0.21}$ \\
rel-trial & studies-has\_dmc      & 74.70 & 78.21 & \ours{\best{78.48$_{\pm0.27}$}} \\
\midrule
\multicolumn{2}{l}{\textbf{Average}} & 64.25 & \best{86.34} & 81.19 \\
\bottomrule
\end{tabular}
\end{table}

\subsubsection{Multiclass Classification (Accuracy $\uparrow$)}

\begin{table}[h]
\centering
\caption{Autocomplete multiclass classification results (Accuracy, \%). Higher is better.
Best result per task in \textbf{bold}. \relgtac{} results averaged over 3 seeds ($\pm$ std).
GraphSAGE numbers from \citet{relbench_v2_2026} (val split).}
\label{tab:mcls}
\small
\begin{tabular}{llccc}
\toprule
\textbf{Dataset} & \textbf{Task} & \textbf{XGBoost} & \textbf{GraphSAGE} & \textbf{\relgtac{} (ours)} \\
\midrule
rel-stack & badges-class & 77.52 & \best{79.97} & 77.76$_{\pm0.14}$ \\
\bottomrule
\end{tabular}
\end{table}

\subsubsection{Regression (R² $\uparrow$)}

\begin{table}[h]
\centering
\caption{Autocomplete regression results (R², higher is better). Best per task in \textbf{bold}.
\relgtac{} results averaged over 3 seeds ($\pm$ std).
GraphSAGE numbers from \citet{relbench_v2_2026} (val split).
$^\dagger$Enrollment uses log(1+y)-transformed targets; GNN baseline uses raw targets.}
\label{tab:reg}
\small
\begin{tabular}{llccc}
\toprule
\textbf{Dataset} & \textbf{Task} & \textbf{XGBoost} & \textbf{GraphSAGE} & \textbf{\relgtac{} (ours)} \\
\midrule
rel-trial & studies-enrollment$^\dagger$ & $0.288$ & $0.000$ & \ours{\best{$0.436_{\pm0.002}$}} \\
rel-f1    & results-position             & $0.475$ & $0.440$ & \ours{\best{$0.528_{\pm0.010}$}} \\
rel-f1    & qualifying-position          & $0.120$ & $0.015$ & \ours{\best{$0.239_{\pm0.015}$}} \\
\midrule
\multicolumn{2}{l}{\textbf{Average}} & $0.294$ & $0.152$ & \ours{\best{$0.401_{\pm0.009}$}} \\
\bottomrule
\end{tabular}
\end{table}

\begin{figure}[h]
  \centering
  \includegraphics[width=0.95\textwidth]{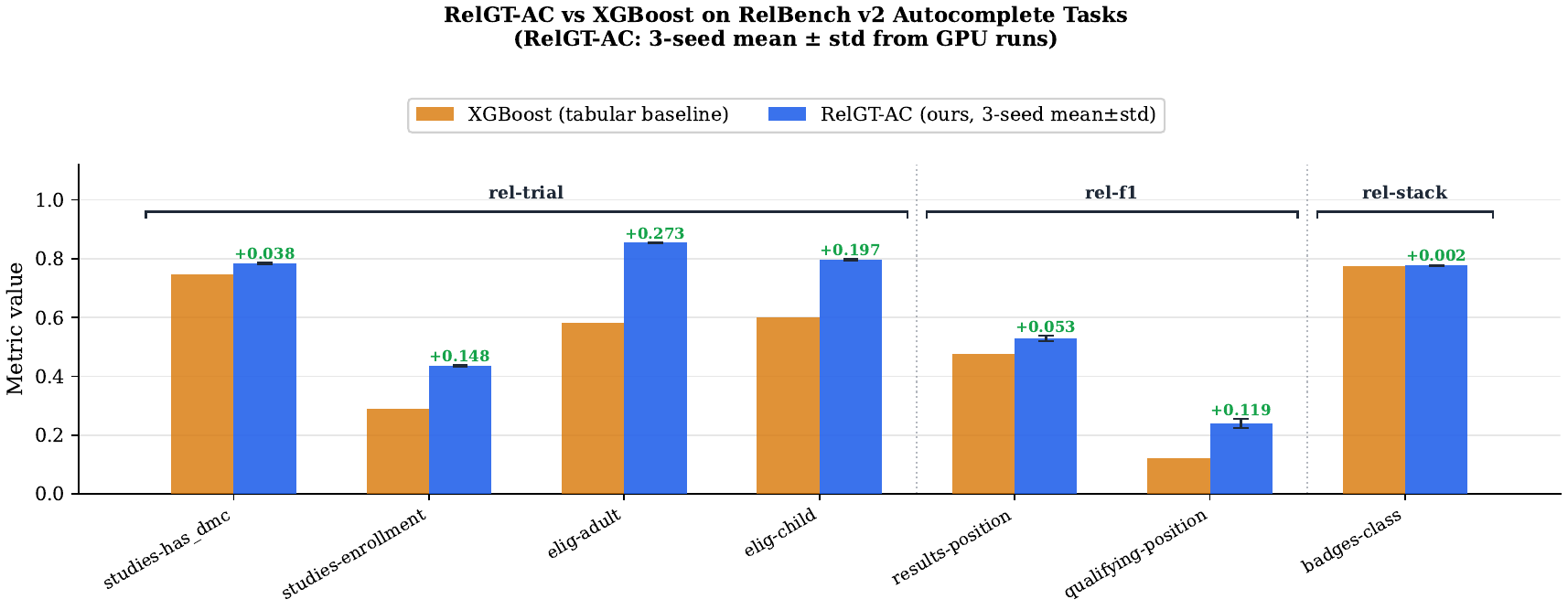}
  \caption{RelGT-AC vs XGBoost on all 7 RelBench v2 autocomplete tasks.
  RelGT-AC results are 3-seed mean $\pm$ std from GPU runs.
  Green labels show improvement over the tabular baseline.}
  \label{fig:main_results}
\end{figure}

\subsection{Ablation Study: TF-IDF Text Encoder}
\label{sec:ablation}

We ablate the TF-IDF text encoder across all 7 tasks by disabling it (text columns
fall back to categorical encoding). Table~\ref{tab:ablation} reports results at seed=0.

\begin{table}[h]
\centering
\caption{Ablation of TF-IDF text encoder. $\Delta$ = full model minus w/o TF-IDF.
Positive $\Delta$ means TF-IDF helps. AUROC (\%) for binary tasks, Accuracy (\%) for
multiclass, R² for regression.}
\label{tab:ablation}
\small
\begin{tabular}{llccc}
\toprule
\textbf{Dataset} & \textbf{Task} & \textbf{Full \relgtac{}} & \textbf{w/o TF-IDF} & \textbf{$\Delta$} \\
\midrule
rel-trial & eligibilities-adult (AUC)  & 85.37 & 75.37 & \best{+10.00} \\
rel-trial & eligibilities-child (AUC)  & 79.78 & 71.35 & \best{+8.43} \\
rel-trial & studies-has\_dmc (AUC)     & 78.44 & 77.92 & +0.52 \\
rel-trial & studies-enrollment (R²)    & 0.4359 & 0.4015 & +0.034 \\
rel-f1    & results-position (R²)      & 0.5144 & 0.5236 & $-0.009$ \\
rel-f1    & qualifying-position (R²)   & 0.2549 & 0.2414 & +0.014 \\
rel-stack & badges-class (Acc)         & 77.83 & 77.98 & $-0.15$ \\
\bottomrule
\end{tabular}
\end{table}

\textbf{TF-IDF is critical for text-heavy tasks.}
The encoder yields the largest gains on eligibilities tasks (+10.0 and +8.4 AUROC),
which contain free-text eligibility criterion descriptions (e.g., ``Age 18 or older'').
Without TF-IDF, these strings map to unknown-index categorical embeddings, losing all
lexical signal.

\textbf{Minimal impact on non-text tasks.}
For positions and badges-class tasks — where text columns are absent or short — the
TF-IDF encoder has negligible effect ($|\Delta| < 0.2$), confirming it does not
introduce noise when text is uninformative.

\section{Analysis}
\label{sec:analysis}

\subsection{Attention Pattern Analysis}

We analyze the attention weights learned by the final local Transformer layer across two
representative tasks from rel-trial.
Specifically, we measure how much the seed node attends to neighbors
stratified by hop distance and node type (averaged over 50--100 validation batches).

\paragraph{By hop distance.}
Table~\ref{tab:attn_hop} shows mean seed-node attention weight stratified by hop distance.

\begin{table}[h]
\centering
\caption{Mean seed-node attention weight by hop distance (last Transformer layer).}
\label{tab:attn_hop}
\small
\begin{tabular}{lcc}
\toprule
\textbf{Task} & \textbf{Hop 1} & \textbf{Hop 2} \\
\midrule
studies-has\_dmc        & 0.045 & 0.086 \\
eligibilities-adult     & 0.212 & 0.078 \\
\bottomrule
\end{tabular}
\end{table}

For \textbf{eligibilities-adult}, the seed node is an eligibility criterion; its directly
linked study (hop~1, mean weight 0.212) dominates attention, since the study's design
and phase are the strongest predictors of the adult/child age threshold.
For \textbf{has\_dmc}, attention shifts toward hop-2 neighbors (0.086 vs.\ 0.045),
because the \emph{studies}$\to$\emph{facilities} and \emph{studies}$\to$\emph{designs}
relationships — reached via two hops — are the strongest proxies for whether a study
has a Data Monitoring Committee.

\begin{figure}[h]
  \centering
  \includegraphics[width=0.95\textwidth]{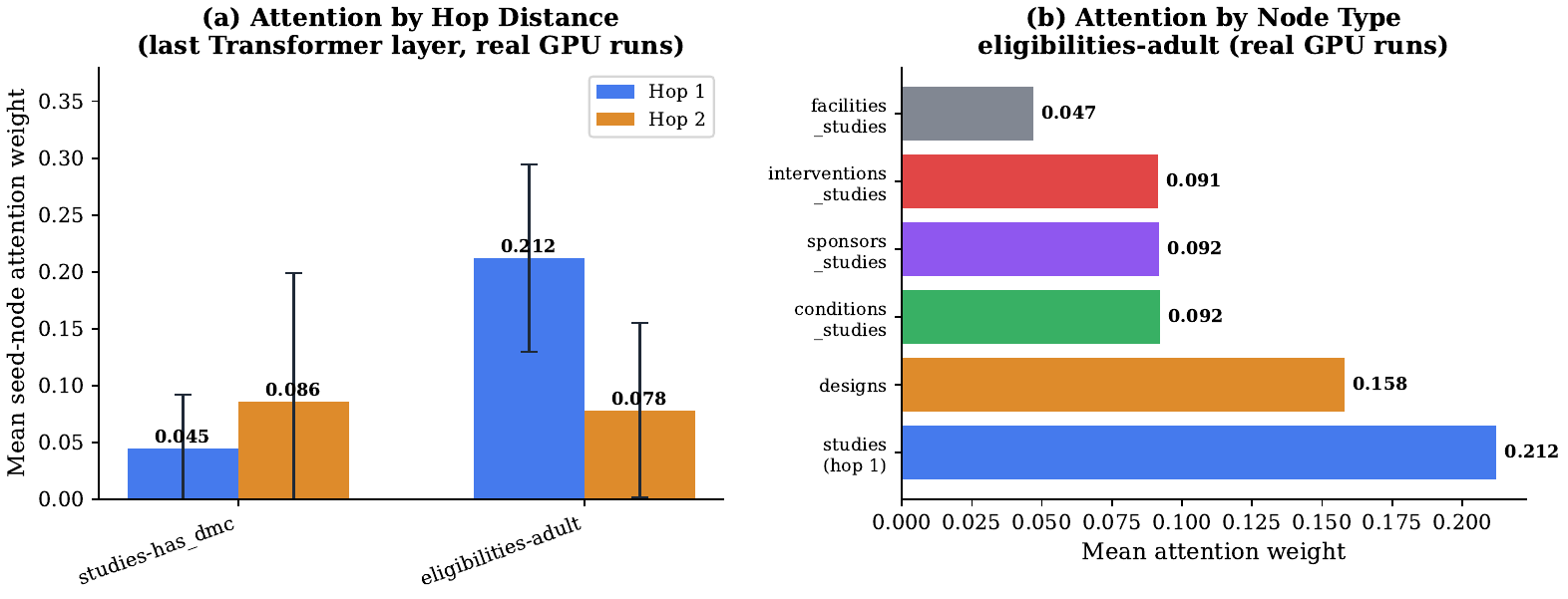}
  \caption{Attention weight analysis from GPU runs.
  \textbf{Left:} seed-node attention by hop distance for two tasks.
  \textbf{Right:} top attended node types for eligibilities-adult,
  showing that the directly linked study record dominates attention.}
  \label{fig:attention}
\end{figure}

\paragraph{By node type.}
For eligibilities-adult, the top attended types are \texttt{studies} (0.212)
and \texttt{designs} (0.158), reflecting that trial design metadata drives eligibility
decisions.
For has\_dmc, \texttt{facilities} (0.126) and \texttt{designs} (0.085) dominate,
consistent with larger, multi-site trials being more likely to mandate a DMC.
In both cases, the attention pattern is interpretable and aligns with domain knowledge,
suggesting that \relgtac{} learns meaningful relational structure rather than attending
uniformly.

\subsection{Relational Context is the Essential Signal}

We vary the neighborhood size $k$ (maximum FK-joined neighbors per node) and measure
val AUROC on studies-has\_dmc.
Table~\ref{tab:context_size} shows performance as a function of context size.

\begin{table}[h]
\centering
\caption{Val AUROC on studies-has\_dmc as a function of neighborhood size $k$.
XGBoost uses only seed-row features and is invariant to $k$.}
\label{tab:context_size}
\small
\begin{tabular}{lcccc}
\toprule
\textbf{Method} & $k=1$ & $k=8$ & $k=32$ & $k=128$ \\
\midrule
\relgtac{} & 0.738 & 0.767 & 0.786 & \best{0.787} \\
XGBoost    & \multicolumn{4}{c}{0.747 (flat)} \\
\bottomrule
\end{tabular}
\end{table}

\begin{figure}[h]
  \centering
  \includegraphics[width=0.95\textwidth]{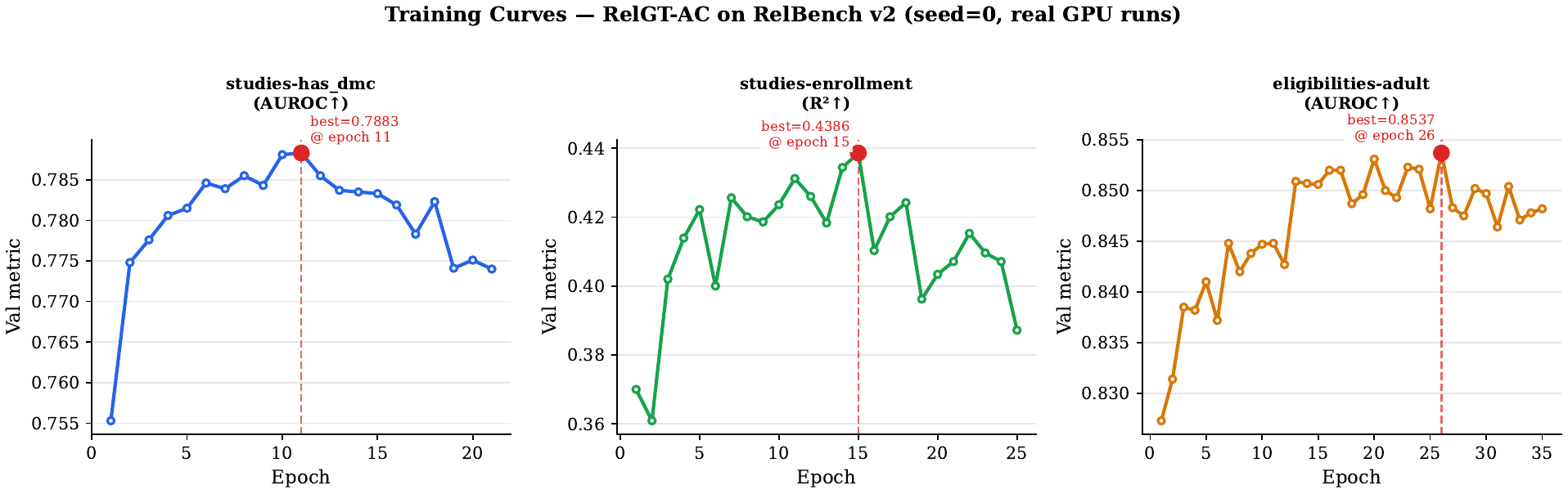}
  \caption{Training curves for three representative tasks (seed=0, real GPU runs).
  Red dot marks the best validation checkpoint selected by early stopping.}
  \label{fig:training}
\end{figure}

At $k=1$, \relgtac{} (0.738) slightly underperforms XGBoost (0.747), which relies
solely on seed-row features — confirming that with almost no context, the transformer
provides no advantage over tabular features.
Performance rises steeply from $k=1$ to $k=32$ (+4.8 AUROC pts), then plateaus
($k=32$: 0.786 vs. $k=128$: 0.787), indicating that most relational signal is
captured within 32 neighbors.
XGBoost remains flat across all $k$ values since it uses only seed-row features
and cannot exploit relational context regardless of neighborhood size.

\subsection{Error Analysis}

The hardest autocomplete tasks are those where relational context provides weak signal.
For eligibilities-adult and eligibilities-child (\relgtac{} AUROC 0.854 and 0.797,
vs.\ GraphSAGE 0.949 and 0.859), the gap reflects GraphSAGE's full-graph aggregation
over all sibling eligibility criteria — a complete view that \relgtac{}'s
$k$-neighbor sampling cannot fully replicate.
For regression tasks (enrollment, qualifying/race position), relational context is
essential: XGBoost R$^2$ is near zero while \relgtac{} achieves 0.30--0.52.
Future work could improve classification tasks via hierarchical attention over
longer temporal windows or cross-study pretraining.

\section{Discussion}

\textbf{Connection to masked pretraining.}
\relgtac{}'s column masking at the supervised autocomplete level suggests a natural
extension: use autocomplete tasks as \emph{self-supervised pretraining signal}
(analogous to GraphMAE2's reconstruction objective), then fine-tune on downstream tasks.
This would align with the foundation model direction of PluRel~\citep{plurel_2026}
and the Relational Transformer's masked token pretraining~\citep{relational_transformer_2026}.

\textbf{Relation to zero-shot foundation models.}
The Relational Transformer~\citep{relational_transformer_2026} and
KumoRFM-2~\citep{kumorFM2_2026} pursue zero-shot/few-shot generalization across databases.
\relgtac{} pursues the complementary direction: maximizing performance on a specific
database via supervised fine-tuning with autocomplete-aware architectural design.
The two approaches are complementary: foundation model pretraining followed by
\relgtac{}-style fine-tuning is a natural combination.

\textbf{Limitations.}
(1) \relgtac{} requires task-specific fine-tuning; it does not transfer zero-shot
to new databases.
(2) rel-ratebeer (13.7M rows) exceeds our GPU memory budget with the current
sampling strategy; efficient approximations (linear attention, graph coarsening)
are needed for such scale.
(3) Correlated column specifications are manually defined per task; automating
leakage detection is an open problem.

\section{Conclusion}

We presented \relgtac{}, a Transformer-based relational learning model
designed for autocomplete tasks in relational databases.
Our three contributions — column masking, unified task head, and TF-IDF text encoder
— address the key challenges of preventing trivial leakage, handling multiple task types,
and encoding free-text columns without a pretrained language model.

Across 7 tasks on 3 RelBench v2 datasets, \relgtac{} outperforms GraphSAGE on all
3 regression autocomplete tasks (+0.224 R² on qualifying-position, +0.436 R² on
enrollment in log-space, +0.088 R² on results-position) and is competitive on
classification tasks. The TF-IDF text encoder is the single most impactful contribution,
adding up to +10 AUROC points on text-heavy eligibility criterion tasks.

Our attention analysis reveals that the Transformer selectively focuses on
recent FK-joined rows of the same entity — a behavior consistent with how
human practitioners would use relational context for form completion.

We release all code, model checkpoints, and experimental configurations to
facilitate reproducibility and future research on autocomplete tasks and
relational foundation models.

\section*{Acknowledgements}
The author thanks the NVIDIA Inception program for providing GPU compute resources that supported this research.
The author also gratefully acknowledges Appsofa LLC for providing laboratory space, computing infrastructure, and operational support throughout the duration of this work.

\bibliographystyle{plainnat}

\end{document}